\documentclass[preprint,journal]{vgtc}       

\pdfoutput=1

\ifpdf
  \pdfoutput=1\relax                   
  \usepackage{graphicx}                
  \DeclareGraphicsExtensions{.pdf,.png,.jpg,.jpeg} 
\else
  \usepackage{graphicx}                
  \DeclareGraphicsExtensions{.eps}     
\fi%

\graphicspath{{figures/}{pictures/}{images/}{./}} 

\usepackage{microtype}                 
\PassOptionsToPackage{warn}{textcomp}  
\usepackage{textcomp}                  
\usepackage{mathptmx}                  
\usepackage{times}                     
\usepackage{cite}                      
\usepackage{tabu}                      
\usepackage{booktabs}                  

\usepackage{url}

\usepackage{amsmath}
\usepackage{algorithm2e}
\usepackage{paralist}
\usepackage{multirow}
\usepackage{makecell}
\usepackage{caption}
\usepackage[usenames, dvipsnames]{color}

\definecolor{angryColor}{RGB}{254, 98, 113}
\definecolor{disgustColor}{RGB}{170, 129, 243}
\definecolor{fearColor}{RGB}{69, 176, 226}
\definecolor{happyColor}{RGB}{251, 202, 53}
\definecolor{neutralColor}{RGB}{189, 189, 189}
\definecolor{sadColor}{RGB}{70, 103, 204}
\definecolor{surpriseColor}{RGB}{62, 200, 69}

\newcommand{\name}{{\textit{EmoCo}}}

\newcommand{\channelview}{{\textcolor{black}{channel coherence view}}}
\newcommand{\channelviewCapital}{{\textcolor{black}{Channel Coherence View}}}
\newcommand{\projectionview}{{\textcolor{black}{sentence clustering view}}}
\newcommand{\projectionviewCapital}{{\textcolor{black}{Sentence Clustering View}}}
\newcommand{\detailview}{{\textcolor{black}{detail view}}}
\newcommand{\detailviewCapital}{{\textcolor{black}{Detail View}}}
\newcommand{\videoview}{{\textcolor{black}{video view}}}
\newcommand{\videoviewCapital}{{\textcolor{black}{Video View}}}
\newcommand{\wordview}{{\textcolor{black}{word view}}}
\newcommand{\wordviewCapital}{{\textcolor{black}{Word View}}}

\onlineid{1102}

\vgtccategory{Research}
\vgtcpapertype{Application/Design Study}

\title{\name: Visual Analysis of Emotion Coherence in \\Presentation Videos}

\author{Haipeng Zeng, Xingbo Wang, Aoyu Wu, Yong Wang, Quan Li, Alex Endert and Huamin Qu}
\authorfooter{
 \vspace{-2mm}
\item
 H. Zeng, X. Wang, A. Wu, Y. Wang and H. Qu are with Hong Kong University of Science and Technology. Y. Wang is the corresponding author. E-mail: \{hzengac, xingbo.wang, awuac, ywangct\}@connect.ust.hk, huamin@cse.ust.hk.
\item 
Q. Li is with AI Group, WeBank, China. E-mail: forrestli@webank.com.
\item
A. Endert is with Georgia Institute of Technology. E-mail: endert@gatech.edu.
}

\shortauthortitle{Biv \MakeLowercase{\textit{et al.}}: Global Illumination for Fun and Profit}

\abstract{
Emotions play a key role in human communication and public presentations. 
Human emotions are usually expressed through multiple modalities.
Therefore, exploring multimodal emotions and their coherence is of great value for understanding emotional expressions in presentations and improving presentation skills.
However, manually watching and studying presentation videos is often tedious and time-consuming. There is a lack of tool support to help conduct an efficient and in-depth multi-level analysis.
Thus, in this paper, we introduce \name, an interactive visual analytics system to facilitate efficient analysis of emotion coherence across facial, text, and audio modalities in presentation videos.
Our visualization system features a \channelview~and a \projectionview~that together enable users to obtain a quick overview of emotion coherence and its temporal evolution.
In addition, a \detailview~and \wordview~enable detailed exploration and comparison from the sentence level and word level, respectively.
We thoroughly evaluate the proposed system and visualization techniques through two usage scenarios based on TED Talk videos and interviews with two domain experts. The results demonstrate the effectiveness of our system in gaining insights into emotion coherence in presentations.
}


\CCScatlist{ 
 \CCScat{K.6.1}{Management of Computing and Information Systems}%
{Project and People Management}{Life Cycle};
 \CCScat{K.7.m}{The Computing Profession}{Miscellaneous}{Ethics}
}

\teaser{
 \includegraphics[width=1\linewidth]{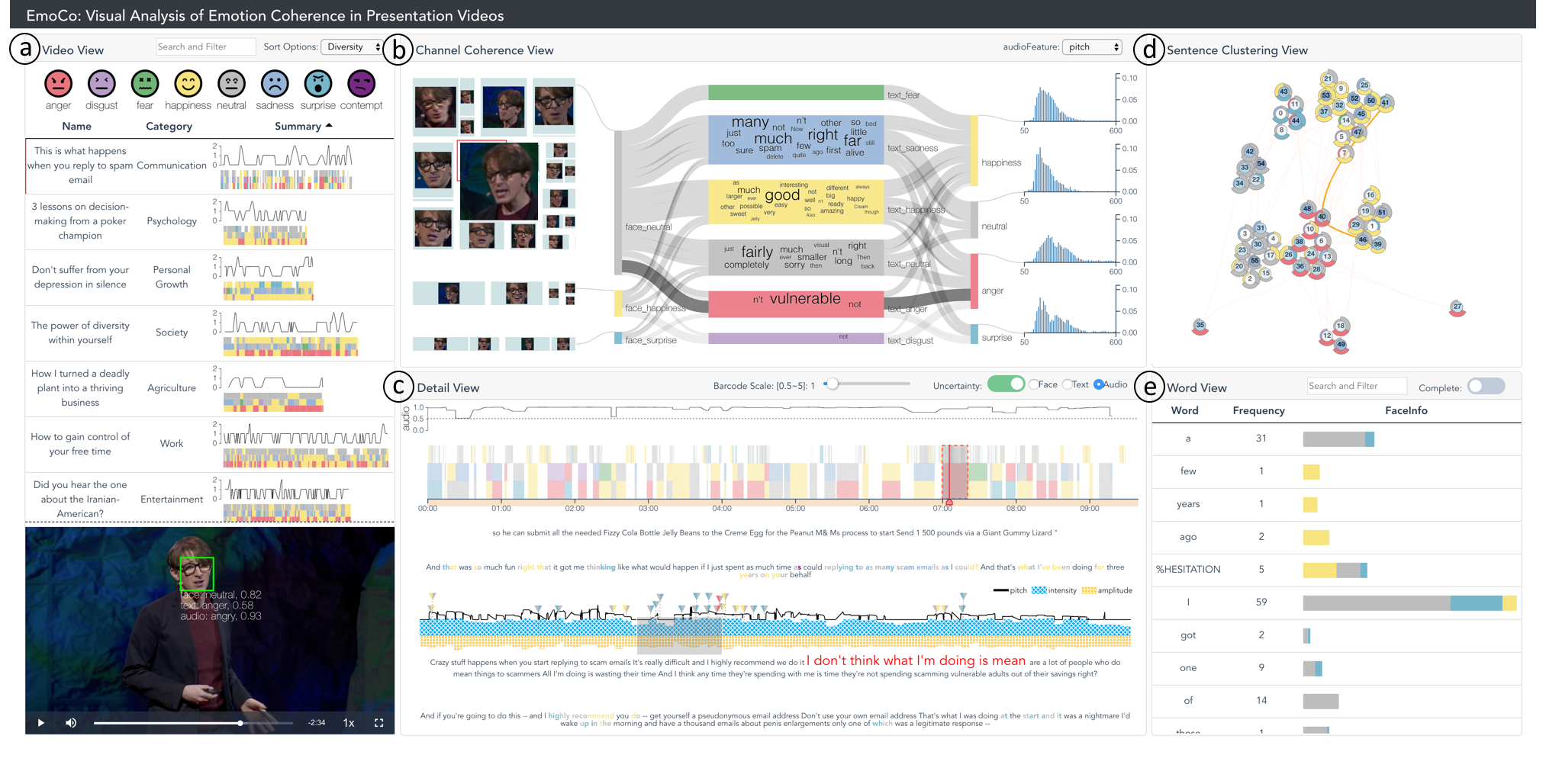}
 \centering
 \vspace{-8mm}
  \caption{Our visualization system supports emotion analysis across three modalities (i.e., face, text, and audio) at different levels of details. The \videoview~(a) summarizes each video in the collection and enables quick identification of videos of interest. The \channelview~(b) shows emotion coherence of the three modalities at the sentence level and provides extracted features for channel exploration. The \detailview~(c) supports detail exploration for a selected sentence with some highlighted features and transition points. The \projectionview~(d) provides a summary of the video and reveals the temporal patterns of emotion information. The \wordview~(e) enables efficient quantitative analysis at the word level in the video transcript.}
\label{fig:systemOverview}
}

\vgtcinsertpkg

\ieeedoi{10.1109/TVCG.2019.2934656}

\begin{document}
\maketitle

\section{Introduction}
Emotions play an important role in human communication and public speaking. Most recent literature advocates using emotional expressions that can improve audience engagement and lead to successful delivery~\cite{gallo2014talk}.
As humans express emotions through multiple behavioral modalities, such as facial and vocal changes, emotion coherence across those modalities can have significant effects on the perception and attitudes of the audience~\cite{darwin1965expression}. Therefore, exploring multimodal emotions and their coherence can be of great value for understanding emotional expressions in presentations and improving skills. Nevertheless, existing research in multimedia~\cite{pfister2010speech, pfister2011real, ramanarayanan2015evaluating} has mainly focused on integrating multimodal features to recognize and analyze the overall emotion in presentations. Thus, they are insufficient for analyzing scenarios with incoherent emotions expressed through each modality, which can occur inadvertently\cite{Reisenzein2013,Yeh2016} or deliberately (e.g., deadpan humor).
To this end, an analysis tool for systematically exploring and interpreting emotion coherence across behavioral modalities is needed to gain deeper insights into emotional expressions.

Visual analytics have been introduced in emotion analysis to ease the exploration of complex and multidimensional emotion data. 
Much effort has focused on analyzing emotions from a single modality such as text data \cite{chen2006visual, oelke2009visual, wu2010opinionseer, zhao2014pearl}, and to a much less extent, videos \cite{tam2011visualization} and audios \cite{chen2008emotion}. While their visualization approaches demonstrate success in analyzing the corresponding emotion modality, it is difficult to integrate them for multimodal analysis due to their different time granularities and dynamic variation. In addition, existing systems for multimodal emotion analysis \cite{zadeh2016multimodal, hu2018multimodal} only encode overall statistics, providing scant support for in-depth analysis, such as identifying dynamic changes of emotion coherence and inferring the underlying emotion states (e.g., deadpan) from videos.
Moreover, those systems do not account for different levels of details, which may result in overlooking important emotion patterns.
In summary, due to the multi-modality and varying granularity of emotional behavior in videos, it is challenging to conduct simultaneous emotion analysis across different modalities and explore the emotion coherence.

To address the above challenges, we work closely with two professional presentation coaches to propose novel and effective visual analytics techniques for analyzing multimodal emotions in presentation videos. 
Following a user-centered design process, we derive a set of visualization tasks based on the interviews and discussions with our two experts. 
As a result, we develop~\name, an interactive visualization system~\textcolor{black}{(Fig.~\ref{fig:systemOverview})} to analyze emotion states and coherence derived from face, text, and audio modalities at three levels of details. The \channelview~summarizes the coherence statistics, and the \projectionview~provides an overview of dynamic emotion changes at a sentence level. Once sentences of interest are selected, the \detailview~enables exploration of emotion states and their temporal variations along with supplementary information, such as voice pitches. Rich interactions are provided to facilitate browsing the videos and inferring the emotion states of the speaker. Two usage scenarios with TED Talk videos and expert interviews demonstrate the effectiveness and usefulness of our approach. 

In summary, the primary contributions of this paper are as follows:

\begin{compactitem}
    \item We design and implement a prototype system to help users explore and compare the emotion coherence of multiple channels, including the emotions from facial expressions, text and audio, with multiple levels of details.

    \item We propose novel designs to facilitate easy exploration of emotion coherence, such as the \channelview~with an augmented Sankey diagram design to support quick exploration of detailed emotion information distribution in each channel, and the \projectionview~based on clustering to help track the temporal evolution.

    \item We present two usage scenarios and conduct semi-structured expert interviews to demonstrate the effectiveness of~\name.
\end{compactitem}

\section{Related Work}
This section presents three relevant topics, namely, emotion modalities, emotion  visualization, and multimedia visual analytics.

\subsection{Emotion Modalities}
 \vspace{-0.7mm}
One central tenet of emotion theories is that emotional expressions involve different modalities, such as facial and vocal behavior \cite{darwin1965expression}. Within this framework, emotion coherence among those channels plays an important role in human communication. Many psychological experiments \cite{weisbuch2010being,tsiourti2019multimodal,muller2011incongruence} have demonstrated the hindering effect of incoherent expressions on emotion perception and recognition by others. Correspondingly, focusing on more modalities than the basic facial expressions alone can enable the discovery of underlying emotion states \cite{aviezer2012body}. Despite such promising benefits, recent psychological research debates that the coherence across emotion modalities is not necessarily high and surprisingly weak for certain types of emotions \cite{fernandez2013emotion, Reisenzein2013}. For example, Reisenzein el al. \cite{Reisenzein2013} found that facial expressions might not co-occur with experienced surprise and disgust. These ongoing debates have motivated our research in multimodal emotion analysis. Specifically, our work looks at how to analyze emotions and their coherence derived from face, text, and audio modalities.

In line with the psychological experimental research, \textcolor{black}{research in affective computing has evolved from the traditional uni-modal perspective to a more complex multimodal perspective \cite{tao2005affective, poria2017review}}.
A great amount of work ~\cite{soleymani2012multimodal, poria2015towards} has focused on utilizing multimodal features to enhance emotion recognition. This line of work has examined different combinations of feature modalities and has identified those that do not contribute to recognition performance. Some work \cite{tzirakis2017end, ranganathan2016multimodal} has employed deep architectures to capture complex relationships among multimodal features; however, they do not explicitly account for their coherence and thus are insufficient in detailed exploration. Our work is different from theirs in two aspects. First, we do not assume that emotions must be coherent on different behavioral modalities, as recent evidence from psychological research claims. Instead, we adopt well-established methods to extract emotions from different modalities and explicitly examine their coherence. Second, we use visual analytics to bring in human expertise in interpreting and analyzing the true emotions from videos, which thus provides a more detailed analysis.

\subsection{Emotion Visualization}
\vspace{-0.7mm}
Emotion visualization has become a prominent topic of research over the last decade. Most effort has focused on analyzing emotions extracted from text data, such as documents \cite{gregory2006user}, social media posts \cite{zhao2014pearl, kempter2014emotionwatch} and online reviews \cite{chen2006visual, oelke2009visual, wu2010opinionseer}. For instance, Zhao et al.~\cite{zhao2014pearl} analyzed personal emotion styles by extracting and visualizing emotion information. On a larger scale, Kempter et al.~\cite{kempter2014emotionwatch} proposed EmotionWatch which summarizes and visualizes public emotional reactions. Less research has addressed facial and audio emotional expressions, which often involve more rapid evolution and smaller time granularity. Tam et al.~\cite{tam2011visualization} utilized parallel coordinates to explore facial expressions in measurement space, supporting the design of analytical algorithms.
However, these systems have mainly centered on uni-modal emotions without considering information from other modalities.
Our visualization approach integrates and tailors those visualization approaches to the varying granularity.

A few systems have been proposed to assist in emotion analysis from a multimodal perspective. Zadeh et al.~\cite{zadeh2016multimodal} utilized histograms to visualize the relationships of sentiment intensity between visual gestures and spoken words in videos. 
More recently, Hu et al.~\cite{hu2018multimodal} inferred latent emotion states from text and images of social media posts and visualized their correlations. 
Nevertheless, their visual approaches only encode overall statistics, lacking an in-depth analysis and visualizations of emotion states and changes at different levels of detail.  
Differing from them, we propose an interactive visual system to help multi-dimensional and multimodal emotion analysis, which few previous systems have addressed.

\subsection{Multimedia Visual Analytics}
 \vspace{-0.7mm}
A number of visual analytics systems have been proposed to assist in video analysis and knowledge discovery. One major challenge is the granularity level, which varies from the video and clip level to the word and frame level. On the one hand, many systems summarize the video content into temporal variables and represent them by line-based \cite{Hoeferlin2013, Meghdadi2013} or tabular charts \cite{Ponceleon2001, Higuchi2017} to enable analysis at the video level. Hierarchical brushing is often introduced to extend the analysis to finer levels \cite{Kurzhals2016}. 
While these systems support an effective visual exploration of the temporal overview and dynamics, they provide scant support for performing analytics tasks such as cluster analysis. On the other hand, some approaches \cite{Matejka2014, Renoust2016} discard the temporal information and consider short clips or frames to be the basic analysis units. For instance, Renoust et al. \cite{Renoust2016} unitized a graph design to visualize the concurrence of people within videos. Our method combines the above-mentioned approaches to support visual analytics at different levels of detail. Moreover, we designed a novel \projectionview~to track the temporal evolution.
Modalities pose another challenge on multimedia mining \cite{Vijayakumar2012, Vijayarani2015}. Exploring the synergy among modalities can reveal higher-level semantic information \cite{Bhatt2011}. While many computational methods have been proposed to support cross-modal analysis, little research has specifically looked at visualization approaches. Stein et al. \cite{Stein2018} proposed a computer vision-based approach to map data abstraction onto the frames of soccer videos. Xie et al. \cite{Xie2018} suggested a co-embedding method to project images and associated semantic keywords on a 2D space. Wu and Qu \cite{Wu2018} proposed a visual analytics system to explore the concurrence of events in visual and linguistic modalities. However, these systems only capture implicit or simple relationships among different modalities, so they are insufficient to promote in-depth analysis. To address those issues, we propose an \emph{enhanced-Sankey view} to explicitly visualize the coherence across three emotion modalities and offer quick exploration of each modality.

\section{Data and Analytical Tasks}
In this section, we first describe the data processing procedures and the derived output. Next, we summarize the tasks based on a user-centered design process with two professional presentation coaches.

\subsection{Data Processing}
\label{sec:dataProcessing}
We conduct a series of data processing steps to extract emotion information from face, text, and audio modalities. We first apply well-established methods to extract information from each modality independently. Next, we fuse those data together based on their semantic meanings and align them at different levels of time granularity.

\textbf{Facial Feature Extraction:} 
The Microsoft Azure Face API~\footnote{https://azure.microsoft.com/en-us/services/cognitive-services/face/} is employed to perform face detection, grouping, authentication and emotion recognition \textcolor{black}{because of its good performance~\cite{khanal2018performance}}. We experimentally consider the preponderant face group to be the speaker, and we greedily merge it with other face groups upon facial authentication, because the same speaker might fall into several groups. The output includes a set of emotions (i.e., \textcolor{black}{anger, disgust, fear, happiness, sadness, surprise, contempt and neutral)} with confidence values for the speaker in each video frame.

\textbf{Text Feature Extraction:} We adopt the predefined, human-labelled text segments from the TED Talk website as the data input, because each one forms a small semantic unit containing a few sentences with similar emotions. \textcolor{black}{The official evaluation of IBM Watson Tone Analyzer service~\footnote{https://www.ibm.com/watson/services/tone-analyzer/} indicates that it performs well on text emotion analysis~\cite{IBMmeasurement}}. Therefore, we use the Tone Analyzer API to extract emotion tones, including \textcolor{black}{anger, disgust, fear, happiness, sadness, and analytical (neutral)}. We mark the last tone as ``neutral'' for consistency.

\textbf{Audio Feature Extraction:} The audios are first segmented in line with the aforementioned transcript segmentation. We use the neural network~\cite{ryokai2018capturing} to filter out audio clips containing laughter, because we observe that they severely affect the emotion recognition results. Next, we compute the Mel Frequency Cepstral Coefficient (MFCC), a feature usually used for audio emotion recognition, from extracted clips. \textcolor{black}{After that, we feed this feature into a baseline model~\cite{barros2016developing}, which achieves a 96\% accuracy in speech emotion recognition for our testing on the RAVDESS dataset~\cite{ravdess}}.
Finally, there are seven detected emotions, including \textcolor{black}{anger, disgust, fear, happiness, sadness, surprise and neutral}.

\textbf{Multi-modal and Multi-level Fusion:}
We fuse the extracted emotion data based on their categories and time granularity.
\textcolor{black}{
For the multimodal emotion categories, since different emotion recognition models are used for each channel and their emotion categories can be different, we use the union of all the possible categories in each modality, thus resulting in eight emotions in total (i.e., anger, disgust, fear, happiness, sadness, surprise, contempt and neutral)}. 
For multi-level fusion, we consider three levels of time granularity (i.e., the sentence level, the word level, and the frame level) based on advice from our two domain experts.
In the previous steps, text and audio emotions have already been aligned at the sentence level, while facial emotions have been extracted frame by frame. To conduct sentence-level fusion, we calculate the most frequent facial emotion in each sentence to represent its predominant emotion. For word-level alignment, since the starting and ending times of each word have been detected by using the IBM Watson Tone Analyzer API, we can easily map the facial, text, and audio emotions to each word based on its detected time period.

\subsection{Data Description}
We collect 30 TED Talk videos\footnote{https://www.ted.com/talks} to explore emotion coherence of presentation videos.
Each video is about 10 minutes long and of high quality, with more than one million online reviews.

After data processing, each TED Talk is described by: 1) the original video and transcript; 2) facial emotions per frame; 3) text and audio emotions per transcript segment; 4) aligned emotions of face, text, and audio modalities per sentence, per word, and per frame. Emotions of each channel are associated with the confidence values output by corresponding models, and further summarized by the preponderant emotion with the highest confidence.

\subsection{Task Analysis}
\label{sec:analyticsTasks}
Following a user-centered design process, we worked closely with two coaches, denoted as E1 and E2, from a presentation training company for about four months. Both coaches have more than five years of experience in presentation training. 
Their current coaching practice is grounded on videotaping presentations to analyze and provide feedback on the performance, which is tedious and time-consuming.
Therefore, we iteratively developed and refined our system to assist them with the video analysis based on their feedback.
Here, we summarize the distilled visualization tasks according to the granularity level as follows:

\vskip 0.1in
\noindent\textbf{Video level exploration} aims to summarize the emotions of each video, and provide video context for detailed exploration: 
\begin{compactitem}
\item [\textbf{T1}]
\textbf{To summarize emotion information in a video.}
It is necessary to summarize emotion information to offer an overview of the entire video collection, which helps users identify videos of interest and thereby guide effective exploration. The emotion information should include the emotion states of each modality and their coherence to represent the overall pattern.

\item [\textbf{T2}] 
\textbf{To provide video context for the analysis.}
Our two domain experts suggest that it is still essential to browse original videos for contextualized exploration in addition to summarized information. Due to the complexity of the data, visualizations should support rapid playback and guided navigation of videos in a screen-space-effective and responsive manner.
\end{compactitem}

\vskip 0.1in
\noindent\textbf{Sentence level exploration} focuses on summarizing emotion coherence of sentences, as well as the detailed information of each sentence:
\begin{compactitem}
\item [\textbf{T3}] \textbf{To summarize emotion coherence across different modalities per sentence.}
Sentences in each transcript segment form a basic semantic unit with the same text emotions in our model. Presenting their coherence with facial and audio emotions is therefore a vital prerequisite for understanding the emotional expressions in presentations. For instance, do speakers' facial expressions react in conformity with a happy message such as jokes?

\item [\textbf{T4}] \textbf{To support rapid location of sentences of interest.}
Our experts are interested in examining how certain emotions are expressed, which demands the ability to rapidly locate sentences with emotions of interest. In addition, they wish to search for sentences with similar emotion expressions in order to comprehend the effects of such behavior on the overall situation. 

\item [\textbf{T5}] \textbf{To display emotion information along with additional features for explanation.}
Our experts suggest to offer additional information, such as the face images, keywords, and prosodic features to verify and better understand the emotion expressions. This information should be displayed with the emotion information to guide the exploration.

\item [\textbf{T6}] \textbf{To show the temporal distribution of emotion states and their coherence.}
The temporal distribution of emotion states and their coherence represents the most detailed and fundamental characteristics. This information should be presented in detail and responsively due to its large scale.
\end{compactitem}

\vskip 0.1in
\noindent\textbf{Word/frame level exploration} shows the emotion of each word/frame, and can reveal changes in how speakers convey their emotions:
\begin{compactitem}
\item [\textbf{T7}] \textbf{To enable the inspection of details of emotion expressions at the word level.}
At a more detailed level, the experts want to explore whether the emotion expressions are associated with words. For instance, are certain kinds of words likely to be accompanied by changes in facial expressions?

\item [\textbf{T8}] \textbf{To reveal transition points of emotion behavior.} 
Our experts are interested in exploring transition between emotion states, because they hope to discover interesting patterns. Therefore, it is important to algorithmically extract transition points and suppress irrelevant details to facilitate a focused analysis.

\end{compactitem}

\section{System Overview}
In this section, we first describe the analytical pipeline, and then introduce each view of the system.

\begin{figure}[!htb]
  \vspace{-4mm}
  \centering
  \includegraphics[width=1\columnwidth]{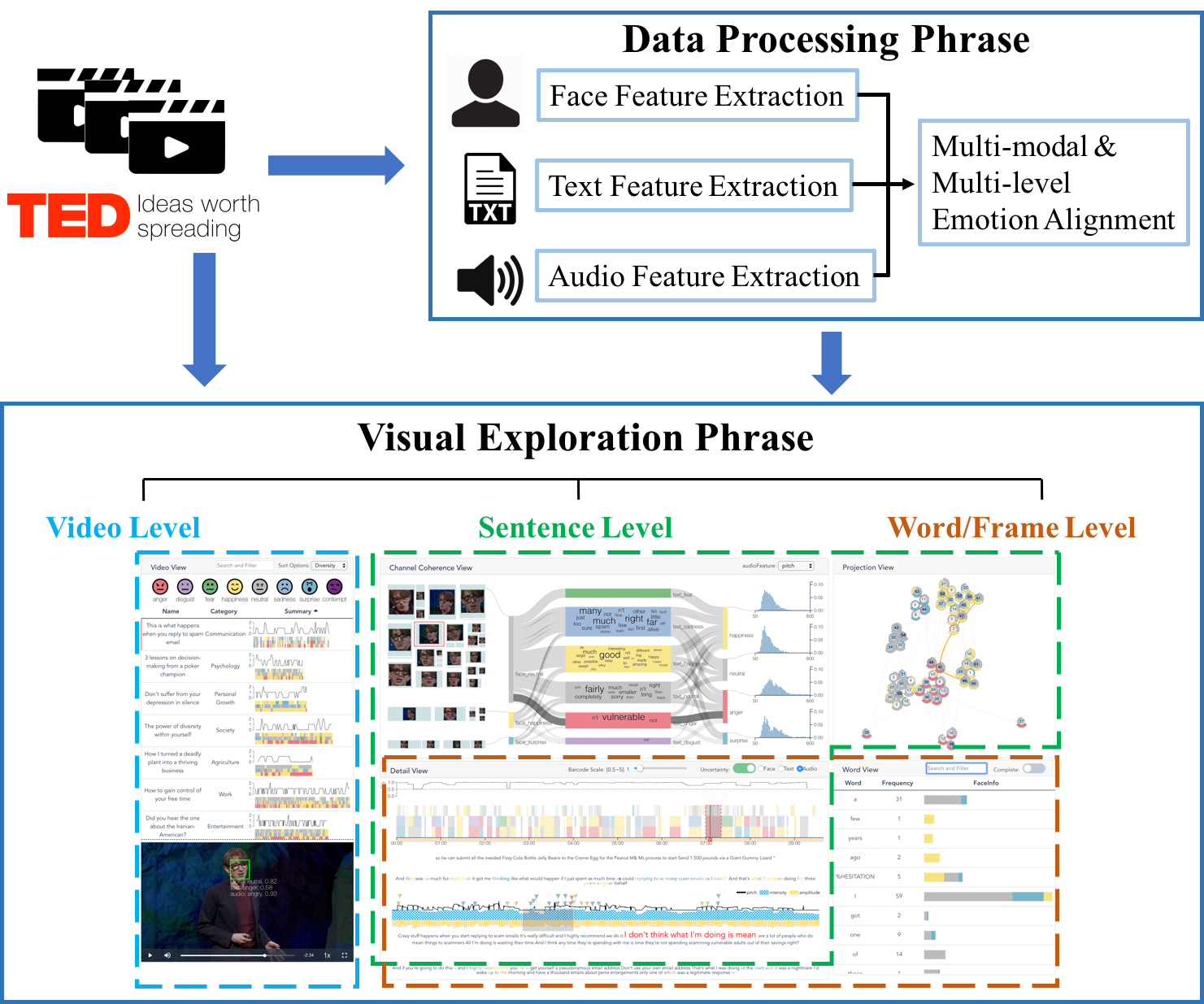}
  \vspace{-5mm}
  \caption{Our visualization system pipeline for multimodal emotion analysis of presentation videos. In the data processing phase, we utilize well-established methods to extract emotion information from different channels. In the visual exploration phase, five coordinated views are provided to support three-level exploration.}
  \label{fig:pipeline}
  \vspace{-3mm}
\end{figure}

As illustrated in Fig.~\ref{fig:pipeline}, our system starts from the data processing phase. After the raw video data is collected, some well-established methods are used to extract emotion information from the face, text and audio channels. This extracted data is stored in MongoDB to facilitate smooth exploration. The data processing phase is explained in detail in Section~\ref{sec:dataProcessing}. In the visual exploration phrase, users can perform three-level exploration with our visualization system. At the video level, users can enjoy a basic overview of each video and select a video of interest for further exploration. Afterward, a summary of emotion coherence based on sentences is provided to help users further explore sentences of interest. Users can then explore some keywords and transition points to further understand the sentences of interest. 

Our system has five views (Fig.~\ref{fig:systemOverview}). The \videoview~(Fig.~\ref{fig:systemOverview}a) presents a list of videos that provides a quick overview of the emotion status of the three channels of each video (T1). Users can easily select a video of interest based on their observation for further exploration. The \videoview~presents the selected video at the bottom to help users directly observe the original information about the video (T2). The \channelview~(Fig.~\ref{fig:systemOverview}b) presents the emotion coherence information of the three channels by using an augmented Sankey diagram design (T3-4). Some corresponding features extracted from different channels are embedded into this view to give some hints on different channels for explanation (T5). The \detailview~(Fig.~\ref{fig:systemOverview}c) presents detailed information of a selected sentence and its contexts to help users analyze a specific sentence (T6-8). The \projectionview~(Fig.~\ref{fig:systemOverview}d) reveals the temporal distribution of emotion similarity across three channels at the sentence level (T6). The \wordview~(Fig.~\ref{fig:systemOverview}e) provides the frequency of each word in the video transcript and allows users to compare different words with the face information and locate specific words in the sentences of a selected video (T7). Some smooth interactions are also provided in the system. For more details, please refer to Section~\ref{sec:interaction}.

\section{Visualization Design}
Based on the analytical tasks mentioned in Section~\ref{sec:analyticsTasks}, we further summarize a set of design rationales with our collaborators to better design our system, which is shown as follows:

\textbf{Multi-level Visual Exploration.}
The mantra ``Overview first, zoom and filter, then details on demand"~\cite{shneiderman2003eyes} has been widely used in exploring complex data. \textcolor{black}{Thus, to explore the data extracted from videos, we follow this mantra in designing our system.}
First, we provide the summary information of the video collection to provide users with some hints that can help them identify a video of interest. After selecting a video, users can further explore the emotion coherence at the sentence level. After selecting a sentence of interest, users can drill down to the word/frame level.

\textbf{Multi-perspective Joint Analysis.}
To facilitate a detailed analysis of emotion coherence from the three channels in videos, various types of information should be provided. For a better interpretation, the features from these channels are extracted and embedded into the corresponding views. Multiple linked views that show different data perspectives are integrated into our proposed system, and users can combine these views to achieve a multi-perspective joint analysis.

\textbf{Interactive Pattern Unfolding.} 
Given that the analysis of emotion coherence in presentation videos contains much hidden knowledge, users need to go through a trial-and-error process. Thus, it is helpful for users to interact with the data directly, so they can observe and interpret the results based on knowledge.

\textcolor{black}{The top of Fig.~\ref{fig:systemOverview}a shows the unified color encoding we adopted. For the five common emotion categories (i.e., anger, disgust, fear, happiness, sadness), we mainly borrow colors from Plutchik’s emotional wheel~\cite{plutchik2001nature} and further carefully design the color mapping for the remaining emotions, where user feedback is also considered.}

\begin{figure}[!t]
  \centering
  \includegraphics[width=0.9\columnwidth]{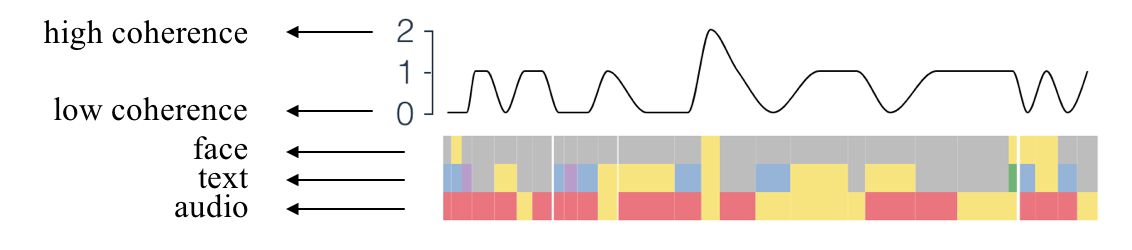}
  \vspace{-3mm}
  \caption{A design for summarizing the emotion information of the three channels of a video. The line at the top explicitly shows the emotion coherence of the three channels. The bar code chart at the bottom shows more details about the exact emotions of each channel.}
  \label{fig:video_view_design}
  \vspace{-3mm}
\end{figure}

\subsection{\videoviewCapital}
\textit{Description}: As shown in Fig.~\ref{fig:systemOverview}a, the \videoview~is divided into three parts. The top part of this view is the legend, which presents our adopted color scheme that helps users know which color is associated with each emotion included in our system. The middle part presents a list of the video information. There are three columns, namely, name, category and summary, and each row provides these three types of information for each video. The first two columns, which indicate the names of videos and their corresponding categories, are easily understandable. The summary column uses a line and a bar code chart to show the coherence of the information of the three channels (Fig.~\ref{fig:video_view_design}), which provides users with a quick overview that helps them in selecting a video of interest (T1). The line is used to explicitly show the degree of \textcolor{black}{emotion coherence} among the three channels. A higher value of the line corresponds to a higher coherence, \textcolor{black}{as shown in Equation 1}. Specifically, ``2'' means that emotions of the three channels are all the same, while ``0'' means that emotions of these channels are all different. 

 \vspace{-4mm}
\textcolor{black}
{
\begin{equation}
D_{coherence} = 
\begin{cases}
    2, & E_{face} = E_{text}~\&~E_{text} = E_{audio},\\
    0, & E_{face} \neq E_{text}~\&~E_{text} \neq E_{audio}~\&~E_{face} \neq E_{audio},\\ 
    1, & others
\end{cases} 
 \vspace{-3mm}
\end{equation}
}

where $D_{coherence}$ indicates the degree of coherence, and $E_{face}$, $E_{text}$ and $E_{audio}$ indicate the emotion types in the corresponding channels.

We also include a bar code chart to show the emotion information of the three channels, with the X-axis representing the length of the video and the Y-axis representing the permutation of the face, text, and audio channels. The color of each rectangle represents the emotions in the three channels. Users can search or filter the video list by typing some keywords in the search function. They can also sort these videos based on specific criteria, such as the coherence, diversity and percentage of one type of emotion. After users click on a row, the video of interest will be selected. 

After a video of interest is selected, the original video is presented in the bottom part of the \videoview~(Fig.~\ref{fig:systemOverview}a) to allow users to explore detailed information (T2). 
Although the extracted information from the video is informative, referring to the original video can sometimes provide a better explanation. In this view, users are allowed to play the video at a slow, normal, or fast speed. After a video is paused, the detected faces are highlighted by green rectangles and the detected emotion information from the three channels will be shown at the same time. When exploring other views, users can easily seek to the corresponding frames by using some provided interactions (Section~\ref{sec:interaction}).

\begin{figure}[!t]
  \centering
  \includegraphics[width=0.99\columnwidth]{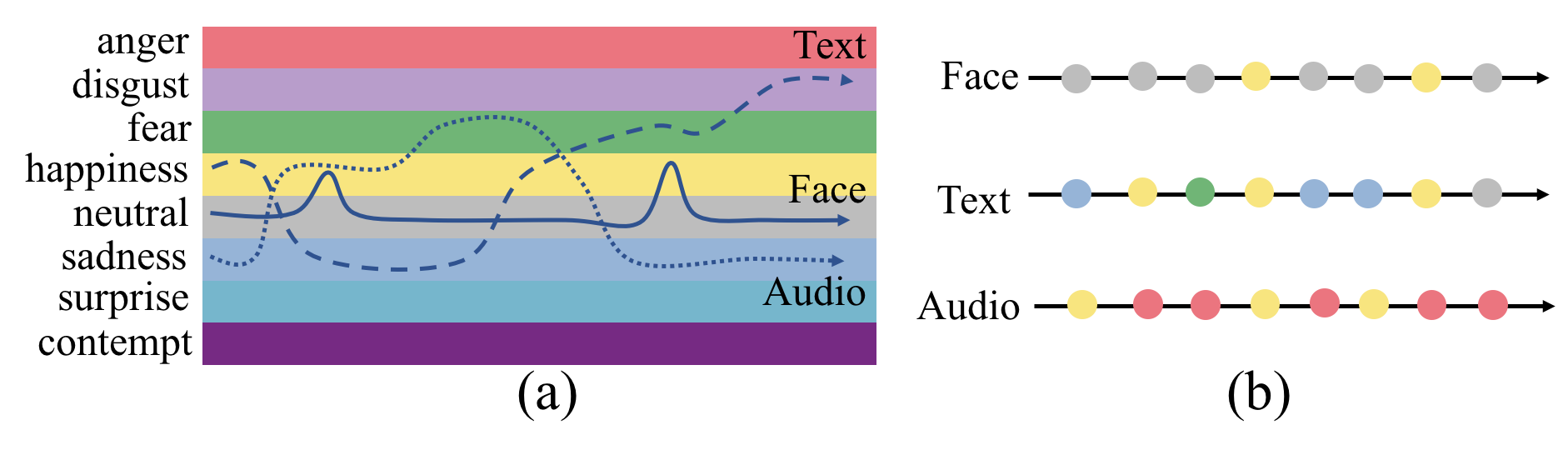}
  \vspace{-5mm}
  \caption{Two alternative designs for summarizing emotion information from the three channels. (a) Three different curves represent the three channels. The position of each curve indicates the current emotion, as the background has eight different-colored emotion bands. (b) Three straight lines represent three channels. Each color dot represents an emotion at one moment.}
  \label{fig:video_view_alternative}
  \vspace{-6mm}
\end{figure}

\textit{Justification}:
Our end users emphasized the need for a quick summary of each video. Originally, we used a scatter plot to visualize a list of videos. Each dot in this scatter plot represented a video, and clusters represented similar videos. However, our end users would like to have more information, such as a quick summary of each video. Thus, they preferred to use a list to show video information. To better show the emotion coherence overview of each video, we considered some alternative designs~(Fig.~\ref{fig:video_view_alternative}). In Fig.~\ref{fig:video_view_alternative}a, the eight different-colored bands in the background represent different emotion categories. The emotion information of each channel was represented by three different curves, allowed us to see how emotions change in each channel. However, this kind of design had obvious visual clutter, so we came up with another design (Fig.~\ref{fig:video_view_alternative}b). We used a straight line to represent each channel, with each color dot indicating an emotion at a specific moment. This design allowed users to easily observe how emotions change on different channels, but it did not make efficient use of space. Therefore, we decided to use a more compact design: a three-row bar code chart. \textcolor{black}{Furthermore, our end users suggested we add a line chart to explicitly show the coherence information and its dynamic trend.} The final design is shown in Fig.~\ref{fig:video_view_design}.

\subsection{\channelviewCapital}
\begin{figure}[!t]
  \centering
  \vspace{-3mm}
  \includegraphics[width=1\columnwidth]{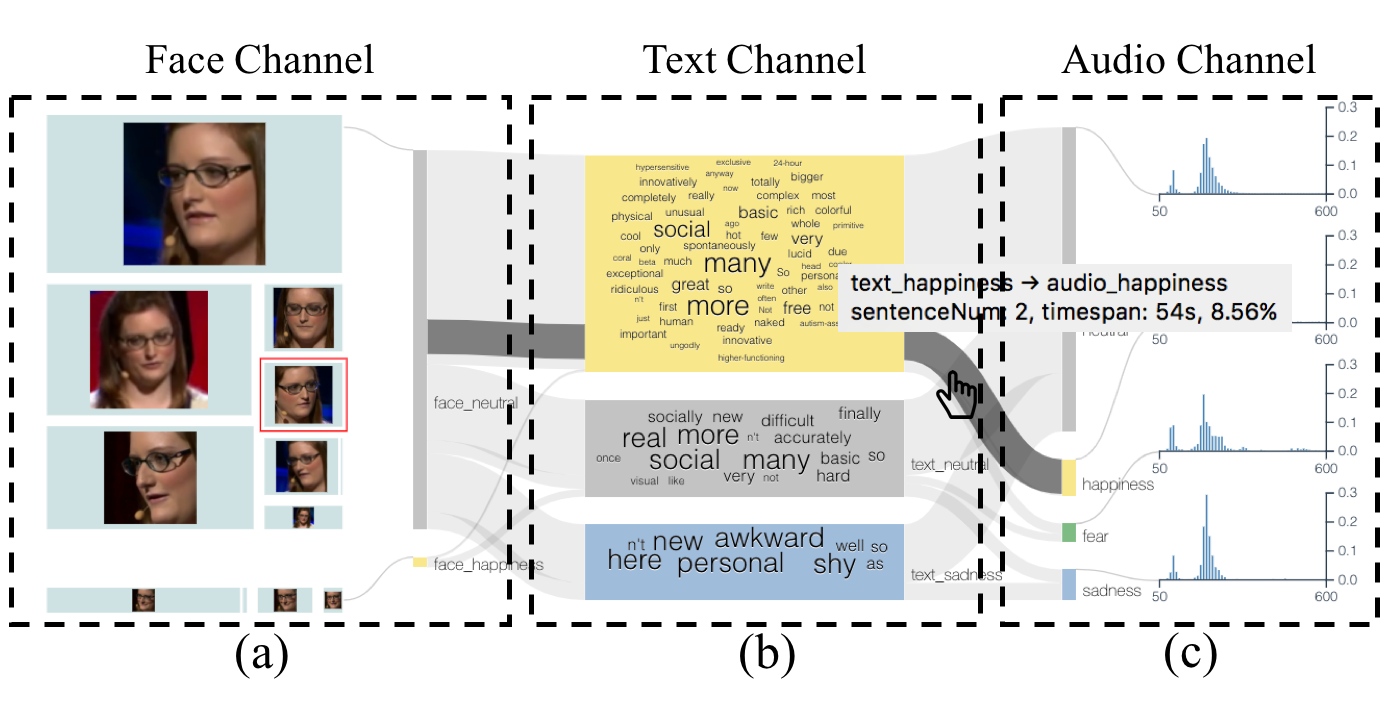}
  \vspace{-9mm}
  \caption{An augmented Sankey diagram design for summarizing emotion coherence from three channels as well as for providing extracted features for explanations. Each node represents one type of emotion and each link represents a collection of sentences with certain emotions shared by two channels, either the face and text channels or the text and audio channels. (a) A treemap-based design to show a quick overview of representative detected faces in the video. (b) A word cloud design to highlight some important words, which gives users some hints about the corresponding context. (c) A histogram design to show the audio feature distribution for different emotions. }
  \label{fig:channel_view_design}
  \vspace{-8mm}
\end{figure}

\vspace{-0.5mm}
\textit{Description}: To show the connection between the three channels in the selected video, as well as showing some features extracted from the corresponding channels, we come up with an augmented Sankey diagram design. As shown in Fig.~\ref{fig:channel_view_design}, this view contains three parts, namely the face channel~(Fig.~\ref{fig:channel_view_design}a) on the left-hand side, the text channel~(Fig.~\ref{fig:channel_view_design}b) in the center and the audio channel~(Fig.~\ref{fig:channel_view_design}c) on the right-hand side. First, we adopt a \textcolor{black}{Sankey diagram design~\cite{schmidt2008sankey}} to visualize the connection among the face, text, and audio channels (T3). The emotion information is detected based on each sentence in the videos. In this way, each node in the Sankey diagram represents one type of emotions, and each link represents a collection of sentences with the emotions between two channels, either face and text channels or text and audio channels. The height of each link represents the total duration of the corresponding sentences. Hence, these links can give users some relevant information on how a speaker conveys his emotions from different channels when he utters these sentences. For example, a link from the left-hand neutral node to the middle happiness node shows that the speaker is talking about something happy while keeping a neutral face, while a link from the middle sadness node to the right-hand neutral node indicates that the speaker is saying something sad in a neutral voice. We add a hover interaction feature to better illustrate the connection among these channels. In Fig.~\ref{fig:channel_view_design}, when users hover over a link between the middle and right-hand nodes, the corresponding link between the left and middle nodes will also be highlighted, thereby highlighting the emotion connection between the three channels.

To provide more information from these channels, we embed features from these channels into the Sankey diagram (T5). 
For each node (face emotion) in the face channel, we adopt a treemap-based design to present an overview of the detected faces, \textcolor{black}{since the data structure is intrinsically hierarchical (i.e., each node contains several links, each link contains several sentences and each sentence contains many face images).}
Each rectangle in treemap represents a cluster (a link), whereas the size of the rectangle represents the number of faces in a specific cluster. As shown in Fig.~\ref{fig:channel_view_design}, the corresponding rectangle area of the link (neutral face, happiness text, and neutral audio) is highlighted. Then, we overlay a representative image on each rectangle. Currently, the representative image for each cluster means the image nearest the center point of the cluster. Other strategies can be easily adopted.
For text information, we embed a word cloud into the middle nodes. After considering their frequency and sentiment, we calculate the importance of each word. Thus, a word cloud is used to show important words in the corresponding sentences and provide users with some context. 
For audio information, we use histograms to visualize the average distribution of corresponding sentences. Users can configure different audio features, including pitch, intensity, and amplitude, and then formulate the corresponding histograms.

\begin{figure}[!t]
  \centering
  \includegraphics[width=0.99\columnwidth]{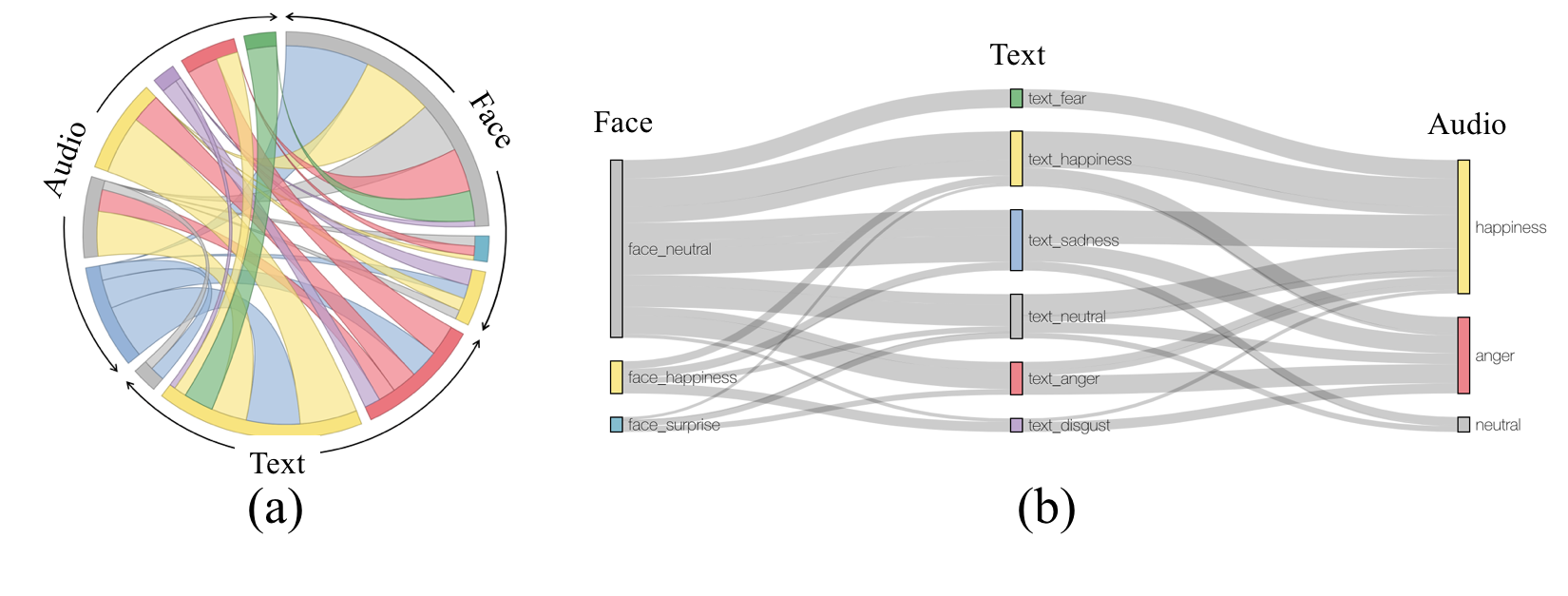}
  \vspace{-7mm}
  \caption{Two alternative designs that were considered for the \channelview. (a) A chord diagram design for showing emotion connection of different channels. Each channel is represented by an arc, and each connection is represented by a link. (b) A normal Sankey diagram design for showing emotion connection between different channels. Emotions in each channel are represented by nodes; their connections are represented by links.}
  \label{fig:channel_view_alternative}
  \vspace{-4mm}
\end{figure}

\textit{Justification}:
To better illustrate the emotion coherence information of different channels, we considered some alternative designs (Fig.~\ref{fig:channel_view_alternative}). At first, we came up with a chord diagram-based design (Fig.~\ref{fig:channel_view_alternative}a), where each channel is represented by an arc, and the links between different arcs represent their connections. Using this design, we could observe emotion coherence information of different channels. However, this kind of design had serious visual clutter in the middle and was not space efficient for embedding extracted features. Therefore, we considered a Sankey diagram design (Fig.~\ref{fig:channel_view_alternative}b). When we presented our prototype to our end users, they commented that it would better to add more information. They also faced difficulty in understanding the connections between different channels. To address this problem, we developed an augmented Sankey diagram design with some interactions, which was favorably received by our end users. 

\subsection{\detailviewCapital}
\vspace{-1mm}
\textit{Description}: As shown in Fig.~\ref{fig:systemOverview}c, the \detailview~consists of two main parts. The bar code chart at the top shows face emotions at frame level and text and audio emotions at sentence level, which provides a more detailed summary than the corresponding bar codes in the \videoview~(Fig.~\ref{fig:systemOverview}a). \textcolor{black}{Users are allowed to adjust the scale of the bar code and scroll along the timeline to explore the details. Due to the imperfect accuracy of emotion recognition, the confidence scores of each channel are encoded as a line chart to give hints about the potential inaccuracy. Also, users can interactively choose to turn on/off the hint or switch to different channels.}
Once users have selected a node or a link in the \channelview~(Fig.~\ref{fig:systemOverview}b), those selected sentences will be highlighted in the bar code chart. Then users are allowed to select a sentence of interest for further exploration. The corresponding sentence context will be shown at the bottom part of the \detailview. Specifically, the sentence being explored is shown in the middle, and the two previous sentences and two following sentences are also shown to provide more context. Three audio features for the selected sentence, i.e., pitch, intensity, and amplitude, are explicitly visualized as a line chart and a theme river, which reveals temporal changes of audio features for the selected sentence. When users brush on part of the sentence, corresponding words will be highlighted. Furthermore, to better visualize the changes of face emotions, we use two inverted right triangles to represent each transition point. The left one represents the emotion before the change, the right one represents the emotion after the change. To avoid visual clutter, dashed lines are used to indicate the location of the transition. Additionally, when transitions happen, corresponding words are also highlighted with colors according to the changes of the facial emotions.

\subsection{\projectionviewCapital}
\begin{figure}[!t]
  \centering
  \includegraphics[width=1\columnwidth]{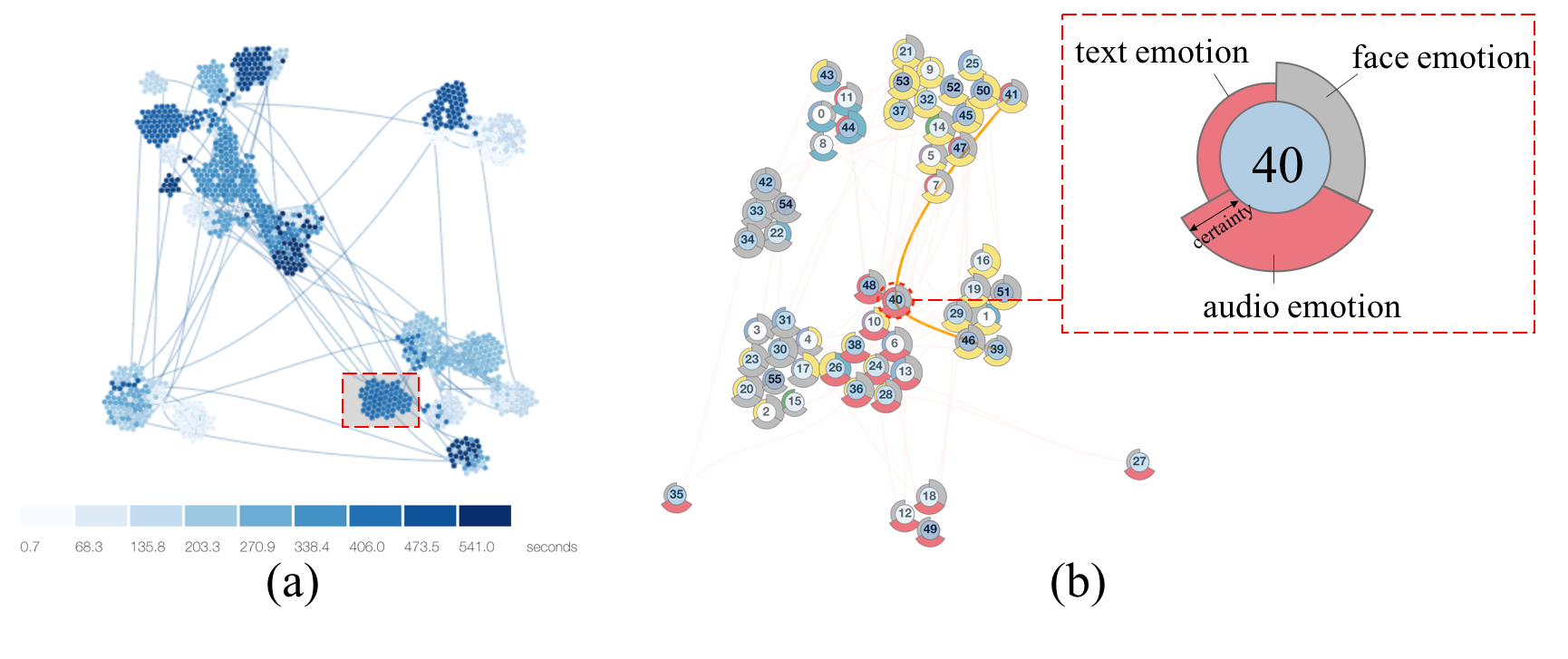}
  \vspace{-9mm}
  \caption{Visual designs for the \projectionview. (a) A frame-based projection without glyph design. (b) A sentence-based projection with glyph design showing emotion information of the three channels, as well as time information.}
  \label{fig:projection_view_design}
  \vspace{-5mm}
\end{figure}

\vspace{-1mm}
To explore how a speaker changes his strategy in conveying emotion over time on different channels, we need to visualize the temporal distribution of the emotion coherence information of different channels. As shown in Fig.~\ref{fig:projection_view_design}b, inspired by the time-curved design~\cite{bach2016time}, we project the emotion information of each sentence as a glyph point on a 2D plane by using the t-SNE projection algorithm, where the vector is constructed as \textcolor{black}{Equation 2}. Points are linked with curves by following time order. To show the information of each sentence more clearly, we design a pie chart-based glyph. 
Three equally divided sectors of a circle are used to encode emotion information of the face, text and audio channels. To be specific, the top left shows text emotion, the top right shows face emotion, and the bottom shows audio emotion. Color is used to represent the type of emotion, and radius is used to represent the emotion probability (certainty). The larger the radius, the higher the emotion probability.
To show temporal information of these sentences, both color and sentence ID in the middle of a glyph are used to represent time order. A lighter color means an earlier time, while a darker color means a later time. 

\vspace{-2mm}
\begin{equation}
Vector = [Pr(E_{face}),~Pr(E_{text}),~Pr(E_{audio})]
 \vspace{-2mm}
\end{equation}
where $Pr(\cdot )$ indicates the detection probability for each emotion in the corresponding emotion category and $E_{(\cdot)}$ indicates one type of emotion in different channels.

\textit{Justification}: Originally, we used a frame-based projection, which projects emotion information from a frame level. As shown in Fig.~\ref{fig:projection_view_design}a, there were too many points and not enough clear information. By observing that those clusters (e.g., the one selected with a red dashed box) of points are almost from the same sentence, our end users commented that there was no need to drill into the frame level in this view. It would be better to explore at the sentence level. Then we considered sentence-based projection, which has a better visual effect. As shown in Fig.~\ref{fig:projection_view_design}b, based on our end users' suggestions, we further embedded glyphs to show both emotion and time information. Our end users were satisfied with this design.

\subsection{\wordviewCapital}
\vspace{-1mm}
Our end users expressed that they would like to further conduct word-level exploration, especially the frequency of the words used and corresponding emotions when uttering these words. In this view (Fig.~\ref{fig:caseTwo}c), we provide detailed information for each word used in the video. Three attributes are shown, namely word, frequency and face information. For each row, the word column directly shows the word used in the video; the frequency column indicates how many times each word is used in the video; and the face information column visualizes the duration of saying this word and the emotion percentage of face emotion by using a stacked bar chart. The length of each component in a stacked bar chart indicates the duration of the expressed type of emotion. For those faces do not be detected, we use dashed areas to represent them (Fig.~\ref{fig:caseTwo}c). For focusing on detected emotions, users are allowed to hide these dashed areas by turning off the switch button. Furthermore, users are allowed to sort the \wordview~by specific criteria, such as frequency, as well as by using a keyword search.

\subsection{Interactions}
\vspace{-1mm}
\label{sec:interaction}
Our system \textcolor{black}{\name}~supports various interactions, empowering users with strong visual analytical abilities. The five views provided in the system are linked together. Here, we summarize the interactions adopted in our proposed system.

\textbf{Clicking}
Once users click a video of interest in the \videoview, the video will be selected and other views will be updated accordingly. In the \channelview, when users click nodes or links of interest, corresponding sentences will be selected and highlighted in the \detailview. In the \detailview, users can click specific sentences to explore their context information. Similarly, users can click a word in the \wordview~to highlight sentences in the \detailview. Furthermore, users are allowed to click the timeline in the \detailview~for seeking corresponding places in the video. 

\textbf{Brushing}
When users brush the bar code in the \detailview~to select corresponding sentences, then corresponding sentences will be highlighted in the \projectionview. Conversely, when users brush some points in the \projectionview, the corresponding sentences will be highlighted in the bar code in the \detailview. In addition, once users select a sentence, they are allowed to brush an area of the selected sentence and identify its words.

\textbf{Searching and Sorting}
In the \videoview~and \detailview, to allow users to quickly discover a row of interest, we add searching and sorting interactions. Users are allowed to search by some keywords and sort the list by a specific criterion.

\section{Usage Scenario}
In this section, we describe two usage scenarios to demonstrate the effectiveness and usefulness of \name~to accomplish the visualization tasks in Section~\ref{sec:analyticsTasks} and discover insights.

\vspace{-1mm}
\subsection{How to be emotional}
\vspace{-1mm}
In this scenario, we describe how Kevin, a professional presentation coach, can find examples for teaching his students to express emotions more effectively.
His teaching is based on the book \textit{Talk like TED} by the keynote speaker Carmine Gallo \cite{gallo2014talk}, where the author attributes the best presentations to be emotional. To strengthen his teaching, Kevin would like to find more examples with considerable emotional expressions. However, it is time-consuming to browse the large video collections and identify representative clips.
Therefore, he \textcolor{black}{refers} to~\name~to explore the videos and find evidence for teaching.

After loading the video collection, Kevin directly notes the video list in the \videoview~(Fig.~\ref{fig:systemOverview}a). 
He wishes to find the video with the most emotions. Thus, he sorts the videos by diversity of emotions, whereby the video entitled \textit{This is what happens when you reply to spam email} appears at the top.
He observes many colors in the corresponding bar code chart, which denotes that this presentation contains diverse emotions (T1). He also notes the frequent fluctuation of its line chart, which indicates that the speaker's emotion coherence varies a lot. As such, he considers this presentation to be representative of good emotional expressions, so he clicks it for further exploration.

To understand overall emotional expressions (T3), he shifts attention to the Sankey diagram in the \channelview~(Fig.~\ref{fig:systemOverview}b). He immediately notices that the three Sankey bars have very different color distributions, and the Sankey links between the same color account for only a small portion of widths. Those suggest that the emotional expressions are incoherent across each modality. 

He decides to explore each modality for detailed understanding. He starts with the leftmost Sankey bar set and finds the predominant grey color, which indicates the most neutral facial expressions. Similarly, he observes a few happy and surprised facial expressions.
Following the face thumbnails to the left, he finds that the speaker has rich facial movements (T5). For example, the speaker tends to raise the corner of his mouth with happy facial expressions, while his mouth tends to open with surprise. As such, Kevin deems facial recognition reliable. In contrast to the leftmost bar set, Kevin observes more emotions, including fear, neutral, happiness, anger, sadness, and disgust, from other two bar sets. He then inspects the histogram to its right, where he finds that anger and surprise tend to yield higher pitches. He considers these results to be reasonable based on his experience.

Next, Kevin decides to inspect detailed expressions with anger, an unusual emotion in presentations. By examining and comparing Sankey links passing red nodes (anger), he identifies the largest link, which connects anger in text and audio modalities with neutral facial expressions. Upon clicking that link, one corresponding sentence is highlighted in the bar code view (Fig.~\ref{fig:systemOverview}c). He selects the sentence to unfold its details.
Following the line chart in the middle, Kevin notices fluctuations of the black line and many glyphs, which denote rapid evolution of voice pitches and facial expressions (T8).
By browsing the video clip, Kevin understands that the speaker expresses an angry message that replying to scam emails is not mean (T4). He emotes and performs theatrical facial and audio expressions, which render his presentation engaging.
Next, he returns back to the bar code view to analyze its context. He notes that both the previous and next sentences have different emotions from the current sentence. Kevin is quite curious. How can the speaker convey various emotions within such a short period? 

He observes a gap between those two sentences, and further finds that the bar code tends to be discontinuous (Fig.~\ref{fig:systemOverview}c). Similarly, he notices large distances between two consecutive sentences in the \projectionview~(Fig.~\ref{fig:projection_view_design}b), which indicates rapid changes of the emotions (T6). Interestingly, he finds that the facial modality behaves quite differently from the other two. Facial information usually does not accompany with text and audio information, and vice versa. To find out what happens in the video, Kevin quickly navigates to those discontinuous parts in the bar code (T2). Finally, he finds that the absence of text and audio information is likely the speaker's presentation style. The speaker usually pauses for a while to wait for the audience's reaction, which is a kind of audience interaction strategy.

Overall, Kevin considers this video to demonstrate the emoting presentation style, which is a good example for his teaching. The speaker adopts a rich set of emotions and tends to be incoherent in a theatrical manner, which renders his presentation infectious and engaging. Upon changing emotions, the speaker often pauses for a while for the audience to react and interact with him.

\vspace{-1mm}
\subsection{How to tell jokes}
\vspace{-1mm}
In this scenario, Edward, another presentation coach, wants to teach a student to incorporate humor in presentations. 
As the student mainly adopts neutral facial expressions, Edward would like to find examples where joke-telling is accompanied by neutral facial expressions to promote personalized learning.

\begin{figure*}[!htb]
  \centering
  \includegraphics[width=2\columnwidth]{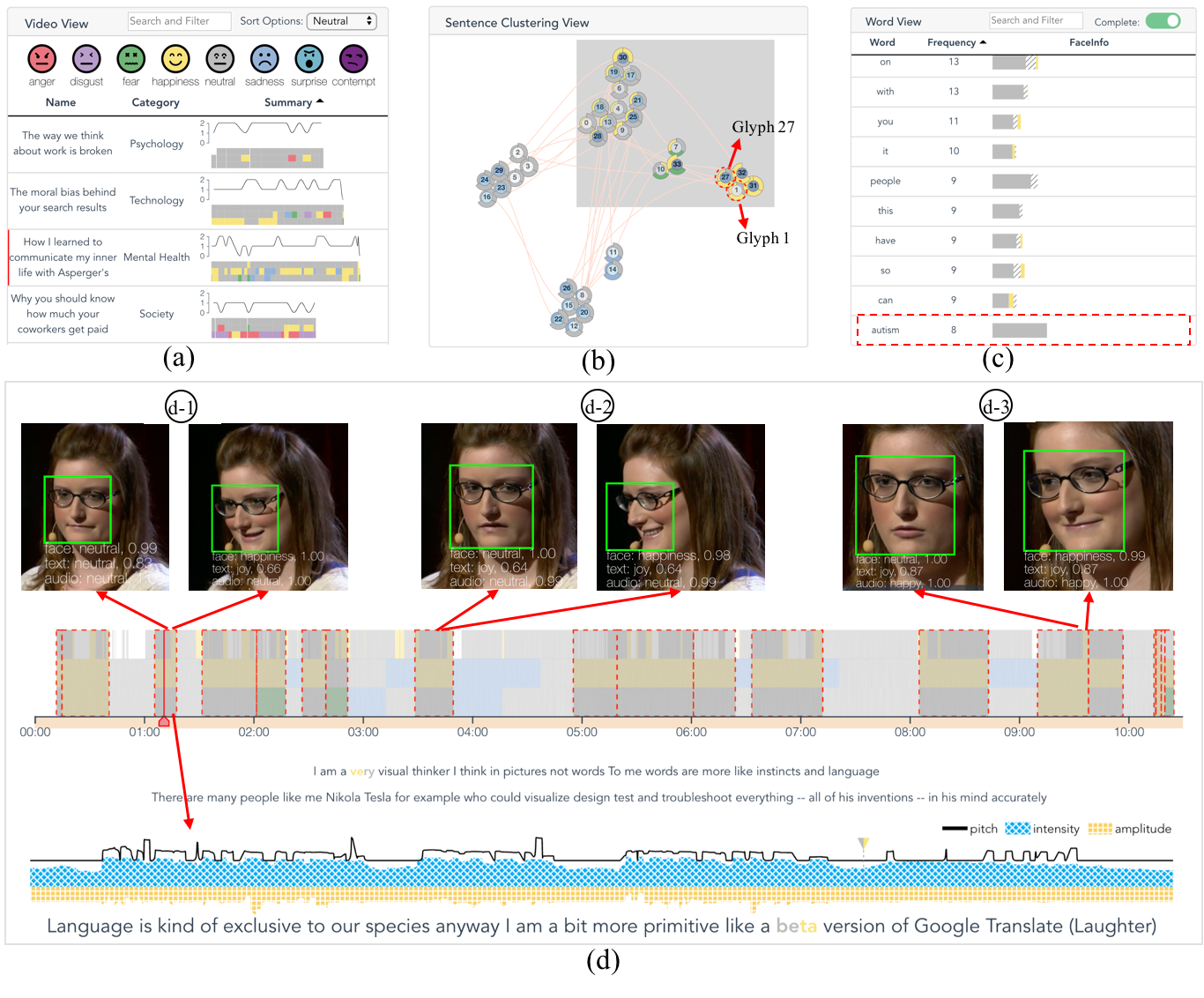}
  \vspace{-6mm}
  \caption{A deadpan humor presentation style. The speaker lacks emotions most of the time when saying something interesting. (a) Select a video with many neutral facial emotions and happy text content. (b) Click the link with happy text and audio emotions in the \channelview~(Fig.~\ref{fig:channel_view_design}), then two sentences represented by the Glyph 1 and Glyph 27 are highlighted. Further, brush an area to find out similar sentences with the highlighted sentences. (c) Sort the \wordview~by frequency to find out which words the speaker tends to use. (d) Some sentences are highlighted with red dashed rectangles in the bar code due to the brush interaction in (b). \textcolor{black}{In particular, we examine three example sentences with neutral facial emotions at first and happy emotions at the end.} The detailed context of the first example sentence is shown.}
  \label{fig:caseTwo}
  \vspace{-2mm}
\end{figure*}

After loading the video collection, Edward sorts them by the percentage of neutral emotions in descending order. By comparing the bar codes in the \textit{Summary} column (Fig.~\ref{fig:caseTwo}a), he finds that the video named ``How I learned to communicate my inner life with Asperger's'' contains preponderant yellow grids in the middle row, which implies predominant happy emotions in the text modality. Thus, he feels interested in this video and clicks it to inspect details in other views.

From the \channelview~(Fig.~\ref{fig:channel_view_design}), he first observes few emotions in each channel where the neutral expressions predominate. 
As highlighted in the darker Sankey link in Fig.~\ref{fig:channel_view_design}, the speaker tends to deliver happy messages with neutral facial and audio emotions.
Since Edward wants to find examples of telling jokes, he clicks on the Sankey link between happy text and audio emotions. Corresponding sentences (Glyph 1 and 27) are highlighted in the \projectionview~(Fig.~\ref{fig:caseTwo}b). To find out other sentences with a similar way of emotion expression, he simply brushes the nearby area of the highlighted glyphs (T6) to locate them in the bar code (Fig.~\ref{fig:caseTwo}d). He then would like to explore how the speaker delivers those happy messages in detail. 

For further exploration, he clicks some of these sentences, and then observes the context or seeks back to the original video (T2). After examining these sentences, he finds that the speaker indeed tells some jokes with a certain presentation style. For example, for the first sentence in the bar code (Fig.~\ref{fig:caseTwo}d-1); the content is shown below. The pitch line tends to be flat (T5), and there are almost no face transition points (T8), which indicates that the speaker does not have many audio changes or face changes when saying this sentence. The speaker in this sentence tells the audience that she is a very visual thinker and not good at language, just like a beta version of Google Translate. After hearing this, the audience laugh. The speaker smiles at the end. As for the second sentence (Fig.~\ref{fig:caseTwo}d-2), the speaker tells the audience that she refused to shower due to hypersensitivity and now she assures her hygiene routine is up to standards. At that moment, the audience laugh and she smiles again. As for the third sentence (Fig.~\ref{fig:caseTwo}d-3), the speaker tells the audience that she loves lucid dreaming, because she can do whatever she wants. Then the speaker throws an example, ``I'm making out with Brad Pitt and Angelina is totally cool with it.'' The audience feel it is very funny and laugh. The speaker also grins. 
Finally, Edward realizes that this is her presentation style in this video. The speaker tells something funny or ridiculous without showing too many emotions. 
In addition, Edward wants to check the words that the speaker uses in the video, which may give him more hints (T7). So he directly sorts the words by frequency in the \wordview~(Fig.~\ref{fig:caseTwo}c). He finds that most of the words the speaker uses are some general words, such as ``you'', ``this'', and ``have''.  \textcolor{black}{Interestingly, he finds that even when the speaker says the bad word ``autism", her facial expressions are neutral, as shown in Fig.~\ref{fig:caseTwo}c, which corresponds to the previous findings}. The speaker does not show too many emotions in the face and audio channels most of the time. From her facial expressions, the audience may feel that the presentation is dry. However, by combing emotion in the other two channels, she makes her presentation very interesting.

Overall, Edward thinks this is a good example for his student to learn from. He thinks the presentation style of this video is deadpan humor, a form of comedic delivery to contrast with the ridiculousness of the subject matter without expressing too many emotions.

\begin{table*}[ht]
\caption{Questions for user interviews.}
\vspace{3mm}
\centering
\begin{tabular}{p{0.1\linewidth}p{0.2\linewidth}p{0.6\linewidth}}
\hline
\textbf{\#}  & \textbf{Aim} & \textbf{Question}\\
\hline
Q1 & Visual Design & It is easy/hard to learn to read the \videoview? why?\\
Q2 & Visual Design & Is it easy/hard to learn to read the \channelview? Why?\\
Q3 & Visual Design & It is easy/hard to learn to read the \detailview? why?\\
Q4 & Visual Design & It is easy/hard to learn to read the \projectionview? why?\\
Q5 & Visual Design & It is easy/hard to learn to read the \wordview? why?\\
Q6 & Interaction Design & It is easy/hard to find a video of interest for further exploration? Why?\\
Q7 & Interaction Design & It is easy/hard to identify sentences/words of interest? Why?\\
Q8 & Interaction Design & It is easy/hard to find similar presentation styles in a video? Why?\\
Q9 & General & Which part of the visual interface do you think can be further improved? How?\\
Q10 & General & Do you think the system is informative for exploring presentation videos?\\
\hline
\end{tabular}
\label{tb:questions}
\vspace{-3mm}
\end{table*}

\section{Expert Interview}
To further evaluate our system, we conducted semi-structured interviews with our aforementioned collaborating domain experts (E1, E2). 
Both E1 and E2 were familiar with basic visualization techniques, such as bar charts and curves. The interviews, which were separated in two sessions, were guided by the set of questions shown in Table~\ref{tb:questions}. The experts were allowed to provide open-ended answers. Each interview session lasted about an hour. After we introduced them to the functions, visual encodings, and basic views of our system, the experts were allowed to freely explore our system in a think-aloud protocol for twenty minutes. In general, we received positive feedback from the experts towards~\name. Both experts appreciated the idea of leveraging visual analytics techniques to support an interactive exploration of presentation videos with novel encodings.

\textbf{Interactive Visual Design.} Both experts confirmed that the system is well designed and follows a three-level exploration hierarchy. The experts were satisfied with the system's views, especially the video, channel coherence, detail, and word views, sharing that ``they are easy to follow and understand.'' As for the \projectionview, E1 commented, ``It is much better than the original frame-based projection. It is easy to understand, but it might not be easy for a novice to use it.'' E1 added, ``Actually, it takes me some time to digest the information presented in the \projectionview, but after practicing more, I find it very useful for finding those similar moments in a presentation video.'' Both experts commented that they could easily find a video of interest by just navigating through the concise summary of each video presented in the \videoview. E1 shared, ``The quick summary provides me with visual hints about the video style, which greatly help me find the video I am interested in.'' Meanwhile, E2 also mentioned that the interactions in the \videoview,~including searching and sorting, were very useful when exploring the presentation videos. Both experts appreciated the ability of our system to identify sentences and words of interest. E2 commented that the current system provides detailed information which facilitates detecting abnormal emotion coherence and emotion changes. ``Usually, I tend to pay attention to those unexpected sentences, such as saying something sad with happy emotion, and I would double check whether it is the real case or caused by some problematic emotion detection. These views are very helpful for me to do this checking.'' E1 was more interested in those emotion transition points. ``These transition points usually indicate different contents in a talk. The \wordview~shows the keywords in the context, allowing the speaker to understand how to improve his presentation skills by using the appropriate words.''

\textbf{Applicability and Improvements.} Both domain experts expressed their interest in applying~\name~to deal with practical problems in their daily work. Previously, to help people improve their presentation skills, they would ask the speakers to conduct multiple practice talks and record them for later analysis. This process is time-consuming and cannot provide any quantitative analysis. E1 stated that ``\name~is the first computer-aided visualization system for emotion analysis and presentation training that I have ever seen, and it can definitely help me analyze those presentation practice videos and train speakers with clear visual evidence.'' E2 especially appreciated the human-in-the-loop analysis by sharing that ``the cooperative process with~\name~is of great benefit in emotion analysis, as the system provides quantitative measures on the level of emotion coherence, and we can determine whether it makes sense or not.'' Our collaboration with the domain experts also suggested directions for future studies. First, E1 suggested that it would be beneficial to show the performance of the speaker in real-time with a score in the \detailview. E2 pointed out that for facial expressions, he identified a few unusual emotions that may result from the lip movements or facial expressions when saying a specific word. Yet, this is mainly due to the accuracy performance of the underlying facial expression algorithms and is beyond the main contributions of our visual analytics system. He also mentioned that it might be helpful to understand the \projectionview~by adding semantic labels to those clusters. In addition, the \wordview~could be further improved by adding functions for analyzing weak words that are widely used in the start or end of presentations.

\section{Discussion}
In this section, we identify limitations of our system and propose directions for future study.

\textbf{Emotion Coherence.}
\textcolor{black}{Since we use different methods to extract emotions from three modalities (i.e., face, text and audio) and their emotion categories vary, we choose the union of all possible emotion categories as the final emotion categories for our further calculation of emotion coherence.
It works well for most videos, since the common categories (e.g., anger, happiness, sadness, and neutral) from different modalities are often the major types of emotions in the three modalities in most videos.
But it can also bring some negative effects on the coherence calculation in certain situations.
For example, contempt only exists in the face channel, so it will be always considered as incoherent with the other two modalities. 
Meanwhile, the current methods of extracting emotions can easily be replaced with other more advanced methods with the same emotion categories in the future.}
What's more, it is still an open question whether emotions should be kept coherent or not during a presentation. In most cases, emotion coherence should be determined based on the specific topic, the content delivered by the speaker and the adopted presentation style. Motivated by this observation, we designed~\name~to help users understand a speaker's coherent or incoherent emotions in a presentation. However, our system still requires users' manual inspection on the presentation video data. In the future, we plan to provide quantitative measures of emotion coherence and integrate applicable advanced algorithms to facilitate this exploration process.

\textbf{Emotion Recognition.} 
We adopt well-established methods to extract emotions from different channels, which can achieve high accuracy.
\textcolor{black}{
When working on this paper, we conducted preliminary evaluations of the emotion recognition accuracy of the three channels on the two videos used in Section 6. We tested 1000 sampled frames, 29 text segments, and the corresponding audio segments for the first video, and we tested 1000 sampled frames, 55 text segments and the corresponding audio segments for the second video. The accuracy of the first video on the face, text and audio channels was 85.4\%, 79.3\% and 89.6\%, respectively, and the accuracy of the second video on the face, text and audio channels was 95.8\%, 87.3\% and 81.8\%, respectively.
Therefore, the emotion recognition methods we used are acceptable in our system.
}
\textcolor{black}{
Meanwhile, we leveraged the confidence scores and designed a line chart in the detail view to give users hints about possible inaccuracy of the emotion recognition.}
In addition, the current emotion recognition algorithms can be easily replaced, when more advanced emotion recognition algorithms are available. 
Furthermore, our work currently considers only eight emotion categories, which may not always satisfy the requirements of different users and tasks. Therefore, we plan to explore more emotion categories to capture more subtle emotions of speakers during presentations.

\textcolor{black}{\textbf{Generalizability.}}
\textcolor{black}{Our system \name~is proposed for analyzing presentation videos.
Two usage scenarios on TED talk videos with predefined text segments are used to demonstrate the effectiveness of our system. 
However, \name~is not limited to TED talk videos and can be easily extended to other presentation videos by adopting automatic transcription techniques~\cite{speechmatics, googleSpeechToText} to extract the text segments from the presentation videos.}

\section{Conclusion}
In this paper, we propose~\name, an interactive visual analytics system to analyze emotion coherence across different behavioral modalities in presentation videos. Our system comprises of five linked views, which allow users to conduct in-depth exploration of emotions in three levels of detail (i.e., video, sentence and word levels). It integrates well-established visualization techniques and novel designs to support visual analysis of videos.
In particular, we propose an augmented Sankey diagram design for analyzing emotion coherence and the clustering-based projection design for tracking the temporal evolution, facilitating the exploration of multimodal emotions and their relationships within a video.
Two usage scenarios based on TED Talk videos and interviews with two domain experts demonstrate that our system can enable efficient and insightful analysis.

In the future, we plan to extend our system to support analysis of additional modalities, such as hand gestures. Moreover, we plan to incorporate advanced data mining techniques to enhance the analysis.
In addition, 
with the capability of the proposed system in analyzing emotion coherence, it would also be interesting to further explore whether our system can be applied to detailed performance analysis of emotion recognition algorithms to further improve their accuracy. 

\acknowledgments{
The authors would like to thank the anonymous reviewers for their valuable comments. This work is partially supported by a grant under Hong Kong ITF UICP scheme (grant number: UIT/142).
}

\newpage
\bibliographystyle{abbrv}

\bibliography{reference}
\end{document}